\documentclass{article}

\PassOptionsToPackage{numbers, compress}{natbib}

\usepackage[preprint]{neurips_2025}

\usepackage[utf8]{inputenc} 
\usepackage[T1]{fontenc}    
\usepackage{hyperref}       
\usepackage{url}            
\usepackage{booktabs}       
\usepackage{amsfonts}       
\usepackage{IEEEtrantools}  
\usepackage{amssymb}        
\usepackage{verbatim}       
\usepackage{nicefrac}       
\usepackage{microtype}      
\usepackage{xcolor}         
\usepackage{algorithm,algpseudocode,amsmath,booktabs}
\usepackage{enumitem}
\usepackage{utfsym}

\usepackage{graphicx}

\title{NOFT: Test-Time \underline{No}ise \underline{F}ine\underline{t}une via Information Bottleneck for Highly Correlated Asset Creation}

\author{%
  Jia Li\textsuperscript{1}\footnotemark[1] \quad Nan Gao\textsuperscript{1}\footnotemark[1] \quad Huaibo Huang\textsuperscript{2} \quad Ran He\textsuperscript{2} \\
  $^{1}$CASIA \quad $^{2}$NLPR, CASIA
}

\begin{document}
\footnotetext[1]{Equal contribution}

\maketitle
\vspace{-25.pt}
\begin{figure}[htbp]
  \centering
  \makebox[\linewidth]{\includegraphics[width=1\linewidth]{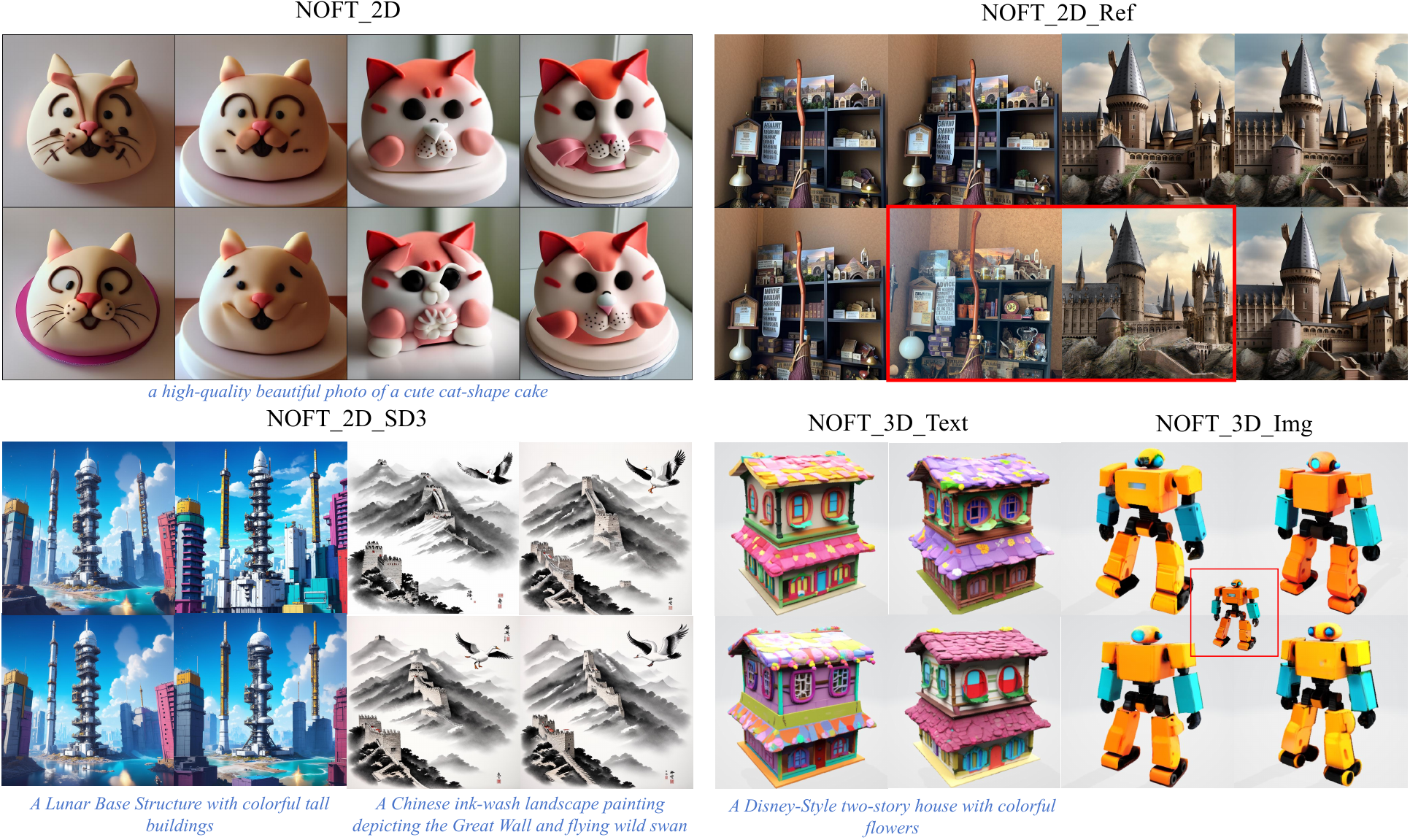}} 
  \vspace{-15.pt}
  \caption{Our method noise finetune (NOFT) completely decouples highly correlated manifold representation learning from dependencies of concept images \cite{word, dreambooth} and external control signals \cite{controlnet, uni}, as well as pre-trained T2I model finetuning \cite{ip, word, controlnet}. Test-time NOFT facilitates high-quality 2D assets \cite{latentdiff, sd3} and 3D assets \cite{trellis} with high contextual fidelity and controllable diversity, under any text or image condition (denoted by red boxes). Zoom in for better observation or go to the Appendix.}
  \label{fig:teaser}
\end{figure}

\begin{abstract}
The diffusion model has provided a strong tool for implementing text-to-image (T2I) and image-to-image (I2I) generation. Recently, topology and texture control are popular explorations, e.g., ControlNet \cite{controlnet}, IP-Adapter \cite{ip}, Ctrl-X \cite{ctrlx}, and DSG \cite{self}. These methods explicitly consider high-fidelity controllable editing based on external signals or diffusion feature manipulations. As for diversity, they directly choose different noise latents. However, the diffused noise is capable of implicitly representing the topological and textural manifold of the corresponding image. Moreover, it's an effective workbench to conduct the trade-off between content preservation and controllable variations. Previous T2I and I2I diffusion works do not explore the information within the compressed contextual latent. In this paper, we first propose a plug-and-play noise finetune \emph{NOFT} module employed by Stable Diffusion to generate highly correlated and diverse images. We fine-tune seed noise or inverse noise through an optimal-transported (OT) information bottleneck (IB) with around only 14K trainable parameters and 10 minutes of training. Our test-time \emph{NOFT} is good at producing high-fidelity image variations considering topology and texture alignments. Comprehensive experiments demonstrate that NOFT is a powerful general reimagine approach to efficiently fine-tune the 2D/3D AIGC assets with text or image guidance.
\end{abstract}

\section{Introduction}
\vspace{-15pt}
\begin{figure}[htbp]
  \centering
  \makebox[\linewidth]{\includegraphics[width=\linewidth]{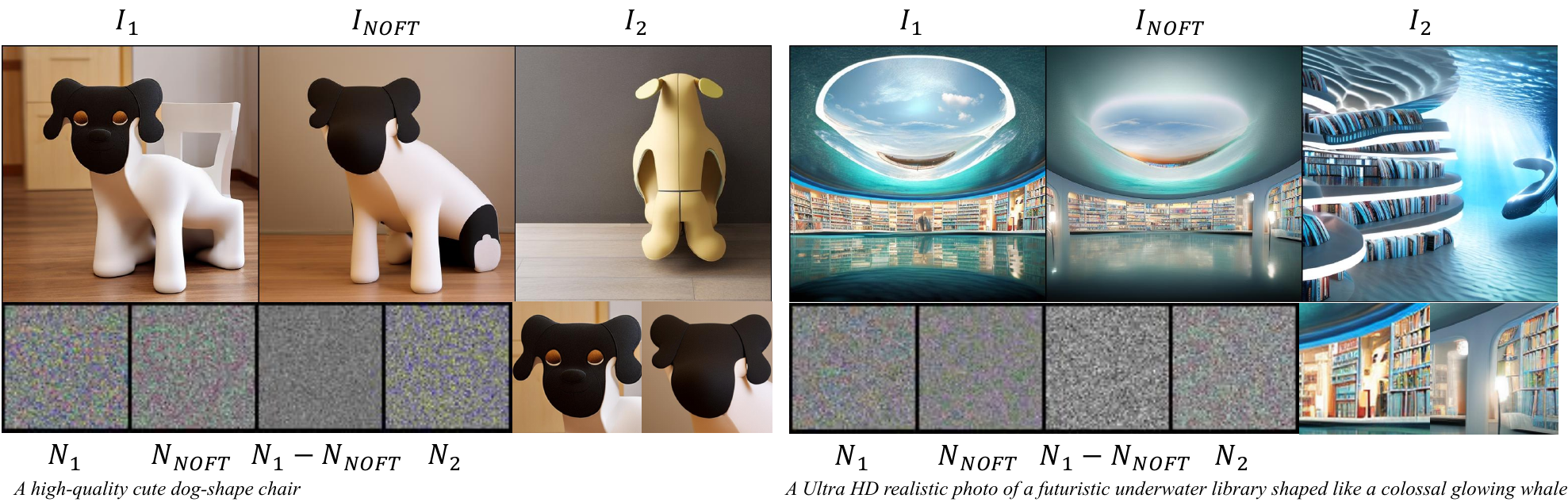}} 
  \vspace{-15.pt}
  \caption{Content-diversity tradeoff: given two distinguished noises, we obtain $N_{NOFT}$ by finetuning $N_{1}$ where adaptively injecting $N_{2}$ based on information bottleneck. The corresponding denoised images are $I_{1}, I_{2}, I_{NOFT}$. As shown by $I_{NOFT}$, the structure and appearance statistics from $I_{1}$ are preserved well, with concurrently improved diversity inherited from the local topological statistic from $I_{2}$.}
  \label{fig:moti}
  \vspace{-5.pt}
\end{figure}
\begin{figure}[htbp]
  \centering
  \makebox[\linewidth]{\includegraphics[width=\linewidth]{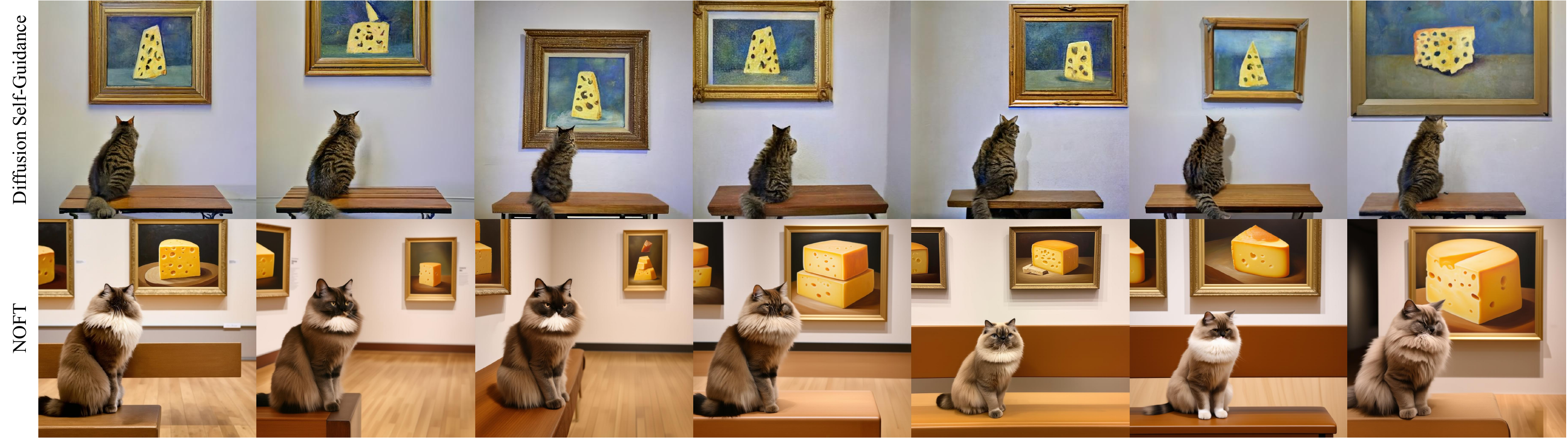}} 
  \vspace{-15.pt}
  \caption{Feature workbench provided by DSG \cite{self} is fine-grained but cumbersome. Our NOFT gives another efficient and diverse workbench to change the properties of objects.}
  \label{fig:self_com}
  \vspace{-8.pt}
\end{figure}

Controllable T2I and I2I are challenging and meaningful tasks for asset creation. Previous diffusion control models try to implement structure or appearance aligned generation explicitly, mainly by feature-level modulation \cite{ctrlx, freecontrol, self}, adapter injection \cite{t2i-adapt, uni, ip}, and model fine-tuning based on external structure or appearance signals \cite{controlnet, word, dreambooth, hyperdreambooth}. On the contrary, we pay attention to the implicit noise-level manipulation on the inherent latent workbench, where we conduct a trade-off of diversity, structure, and appearance simultaneously. While achieving similar editing effects to DSG \cite{self} in Figure \ref{fig:self_com}, our method doesn't require any explicit guidance, e.g., position, size, shape, leveraging implicit noise finetune \emph{NOFT}.

Recently, test-time noise searching \cite{inference} has proved that better noise plays an important role in diffusion performance. To be specific, the noise seems messy, but it implicitly represents a certain context of the image that will be generated based on this noise. Two examples are illustrated in Figure \ref{fig:moti}. Given the same text prompt, different noises, i.e., $N_{1}, N_{2}$, are denoised as corresponding images, i.e., $I_{1}, I_{2}$. Note that $I_{1}$ and $I_{2}$ have respective structures and textures, which demonstrates that Gaussian noise inherently encodes contextual information. 

Furthermore, we fine-tune $N_{1}$ slightly based on our algorithm in the test time of the diffusion model. Concretely, we randomly compress some local information of $N_{1}$ and adaptively inject other information of $N_{2}$ for diversity in an implicit manner, inspired by information bottleneck \cite{ib,iba} and Sinkhorn optimal transport \cite{sinkhorn, otseg}. And then, we obtain the fine-tuned noise $N_{NOFT}$ based on which $I_{NOFT}$ is synthesized. Qualitative results show that $I_{NOFT}$ preserves the global layout and appearance of $I_{1}$, meanwhile exhibiting significant diversity. More remarkably, the local structure manifold pattern from $I_{2}$ is transferred to $I_{NOFT}$. 

Our paper presents several significant contributions, mainly including three folds:

1. We first entirely explore the implicit noise representation rather than other explicit control manners, such as attention matrices \cite{ctrlx, self}, intermediate activations \cite{freecontrol, self}, or external control signals \cite{controlnet, uni, ip, layoutdiffusion, instancediffusion, migc}. Remarkably, test-time noise finetune \emph{NOFT} demands merely brief training while maintaining full disentanglement from the diffusion model's forward and denoising process. Considering information compression and diversity injection, our approach achieves highly correlated 2D/3D results, with any text or image condition. 

2. We present an efficient and effective Optimal-Transported Information Bottleneck (OTIB) module that provides a trade-off between preservation of topology and texture, as well as synthesis variety. Moreover, the proposed Sinkhorn attention further builds up fidelity and quality of asset creation.

3. Our proposed NOFT is capable of being adaptive for multiple asset creation tasks, base architectures, and model checkpoints. Compared with state-of-the-art structure-aligned and appearance-aligned approaches, comprehensive experimental analyses demonstrate that NOFT is the first effective plug-and-play implicit controller for pre-trained T2I models with exceptional context preservation and generation diversity.
\vspace{-10pt}
\section{Related work}
We briefly introduce diffusion control methods, diffusion seed implementation, and information compression works in this section.
\paragraph{Diffusion control.} 
On one hand, pre-trained T2I foundational models \cite{latentdiff} are potentially able to generate diverse images taking advantage of the random noise initialization. On the other hand, uncertainty from the Gaussian noises makes it hard to synthesize credible images with a certain topology or texture. To address this matter, previous diffusion control methods compose different adapters independently \cite{t2i-adapt, uni}, or conduct adaptively feature modulations \cite{controlnet, ctrlx}, and model finetune \cite{word, dreambooth, hyperdreambooth} to facilitate alignment of internal diffusion knowledge and external control signals.

\emph{Topology alignment} SD-based methods have demonstrated strong generalization capabilities and composability while maintaining high creation quality \cite{gligen, uni, reco, spatext, layoutdiffusion, instancediffusion, migc}. External control signals include Canny edge, depth map, human pose, line drawing, HED edge drawing, normal map, segmentation mask (used in \cite{controlnet, uni}), as well as 3d mesh, point cloud, sketch (used in \cite{ctrlx}), etc. FreeControl \cite{freecontrol} manipulates the specific-class linear semantic subspace to employ structural guidance. Semantic signal usually possesses higher freedom than low-level vision signals. Note that our NOFT does not depend on any external structure control signal.

\emph{Texture alignment} methods try to realize I2I by image prior embedding or few-shot weight adaptation. General I2I methods extract global semantic embedding from the referenced images \cite{uni, ip, t2i-adapt}. Personalized model concerning specific concept needs pretrained T2I diffusion finetuning based on a small set of image samples \cite{dreambooth, word, break, orthogonal, hyperdreambooth}. FreeControl \cite{freecontrol} uses intermediate activations as the appearance representation, similar to DSG \cite{self}. However, our NOFT achieves superior appearance alignment performance without personalized concept data or model fine-tuning.
\paragraph{Diffusion seed.}
Previous diffusion control methods only treat Gaussian noise as a flexible random generation seed \cite{controlnet, uni, ip, layoutdiffusion, instancediffusion, migc, dreambooth, word, break, orthogonal, hyperdreambooth}. They constrain the pre-trained diffusion model using external structure or textural data. Nevertheless, some diffusion inversion works \cite{flowprior, ddim, null} show high-fidelity image reconstruction and editing. Seed searching \cite{inference} is beyond the denoising steps for high-quality image generation. These methods establish the critical role of noise representation, which is demonstrated by Figure \ref{fig:moti} as well. Therefore, we explore the implicit structure and appearance alignment based on noise in this paper.
\begin{figure}[htbp]
  \centering
  \makebox[\linewidth]{\includegraphics[width=\linewidth]{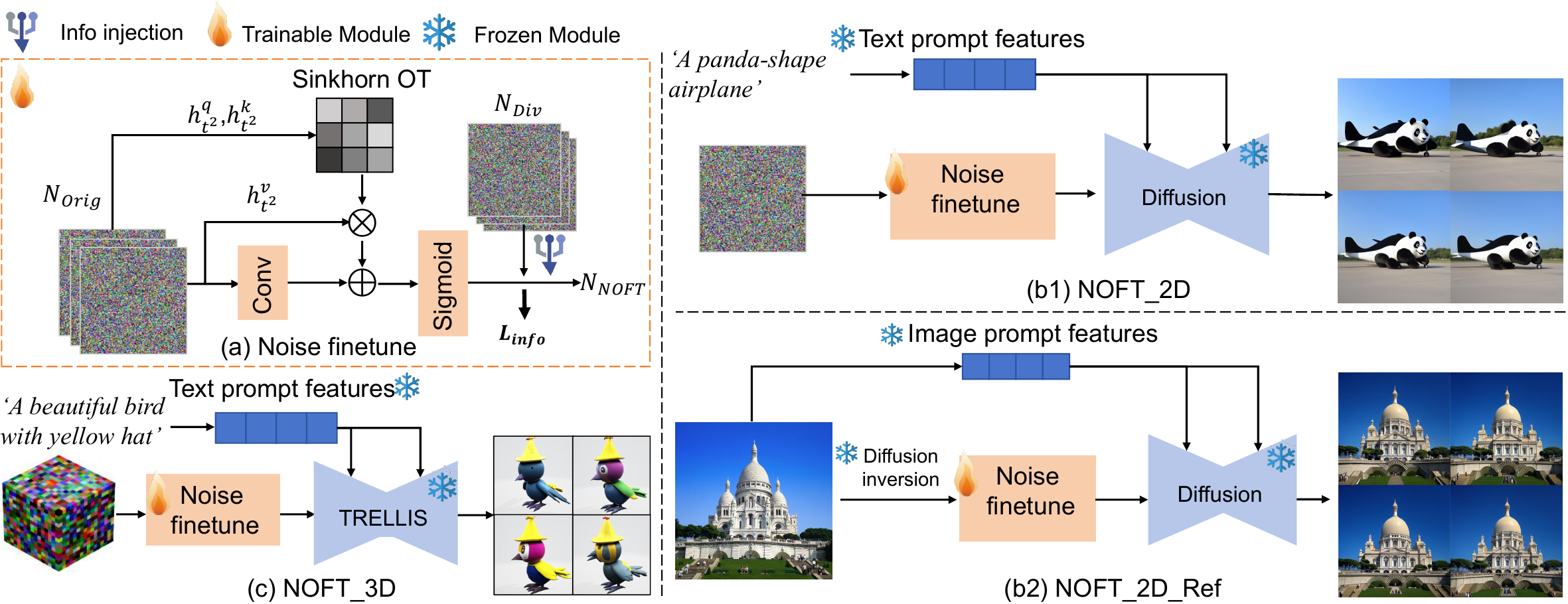}} 
  \vspace{-15.pt}
  \caption{Method overview: as a plug-and-play content controller, NOFT can be employed for 2D/3D generation tasks, different architectures and model checkpoints. NOFT consists of a Sinkhorn Attention module and an information bottleneck module. We obtain $N_{NOFT}$ by information compression of $N_{Orig}$ and information modulation of $N_{Div}$. More details are introduced in Section \ref{Approach}.}
  \label{fig:pipeline}
  \vspace{-10.pt}
\end{figure}
\paragraph{Information bottleneck.}
Information bottleneck (IB) \cite{ib} plays a representation trade-off between information compression and information preservation for neural learning tasks. 
Furthermore, VIB \cite{vib} leverages variational inference to facilitate the IB neural compression. IBA \cite{iba, infoswap} polishes the attribution information based on KL divergence \cite{kl} to effectively disentangle relative and irrelative information concerning the classification task. We will introduce our information bottleneck in Section \ref{Preliminaries}, \ref{Approach}.
\section{Preliminaries}
\label{Preliminaries}
The latent diffusion model \cite{latentdiff} conducts a denoising process on the compressed latent from the Gaussian noise distribution. The distribution regularization of the latent diffusion model is formulated as:
\begin{equation}
\mathcal{L}_{ldm}=\mathbb{E}_{z,c,t,\epsilon}[\Vert\epsilon-\epsilon_{\theta}(z_{t}=\sqrt{\overline{\alpha}_{t}}z+\sqrt{1-\overline{\alpha}_{t}}\epsilon, c, t)\Vert_{2}^{2}],
\end{equation}
where $z$ means the manifold compressed via the encoder of VAE.
$\epsilon \sim \mathcal{N}(0,\mathbb{I})$ has variance $\beta_{t}=1-\alpha_{t}\in(0,1)$ used to conduct noisy manifold reparameterization. The denoised manifold of the pre-trained diffusion model is calculated as follows:
\begin{equation}
\Tilde{z}_{0}=\frac{z_{t}}{\sqrt{\overline{\alpha}_{t}}}-\frac{\sqrt{1-\overline{\alpha}_{t}}\epsilon_{\theta}(z_{t},c,t)}{\sqrt{\overline{\alpha}_{t}}}.
\label{eq3}
\end{equation}
Our method NOFT completely decouples highly correlated noise representation learning from not only the dependencies of concept image \cite{word, dreambooth} and external control signals \cite{controlnet, uni, freecontrol}, but also pre-trained model finetuning \cite{ip, word, controlnet}. We define our noise finetuning as:
\begin{equation}
\label{equ:g}
\theta^{*} =argmin_{\theta}\mathbb{E}_{N_{Orig},N_{Div}}[\mathcal{L}_{noise}(NOFT_{\theta}(N_{Oirg}, N_{Div}),N_{Orig})+\mathcal{L}_{info}(NOFT_{\theta}(N_{Orig}))],
\end{equation}
where $NOFT_{\theta}$ is the generator of NOFT, $N_{Orig}$ is the source noise, and $N_{Div}$ is the random noise for sampling diversity. $\mathcal{L}_{noise}$ aims to provide pixel-level regularization of $N_{Orig}$ for structure and appearance alignment, and $\mathcal{L}_{info}$ explores controlling appropriate neural feature leakage with consideration of contextual preservation.

Let's denote the original input data, the corresponding label, and compressed information by $X$, $Y$, and $Z$. The information compression principle \cite{ib, ib_concept} is a trade-off between information preservation and the minimal sufficient representation supervised by the target signal, by means of maximizing the sharable information of $Z$ and $Y$ while minimizing that of $Z$ and $X$:
\begin{equation}
\label{equ:mib}
\mathop{\max}_{Z} \mathbb{I}(Y;Z) -\beta\mathbb{I}(X;Z),
\end{equation}
where $\mathbb{I}$ means the mutual information and $\beta$ is a trade-off weight.  Let $R$ denote the feature representations of $X$, and the information loss definition of $\mathbb{I}(X;Z)$ is formulated as:
\begin{equation}
\label{equ:3}
\mathbb{I}(X;Z) \triangleq{\mathbb{I}(R;Z)} \triangleq{\mathcal{D}_{KL}[p(Z|R)\Vert q(Z)]},
\end{equation}
where $q(Z)$ with Gaussian distribution is a variational approximation of $p(Z)$ \cite{iba}. $\mathcal{D}_{KL}$ is the KL divergence \cite{kl} used to represent  the distance between two distributions.

\section{Approach}
\label{Approach}
In this section, we provide a detailed introduction to our proposed NOFT method, including the overall pipeline in Section \ref{total}, optimal transport information bottleneck (OTIB) module in Section \ref{otib}, along with the training loss in Section \ref{loss}.
\subsection{Overall pipeline}
\label{total}
As shown in Figure \ref{fig:pipeline}, NOFT can manipulate random noise with text or image conditions in 2D \cite{diffusion, latentdiff, sd3} or 3D data \cite{trellis} distribution.

\subsubsection{NOFT\_2D}
As for none-referenced NOFT\_2D, given a text prompt denoted by $\text{'S'}$, diverse images can be synthesized based on:
\begin{equation}
\label{equ:2D}
I_{NOFT} =G_{\phi}^{2D*}(NOFT_{\theta}^{2D}(N_{Oirg}, N_{Div}), \text{'S'}),
\end{equation}
where $G_{\phi}^{2D*}$ is the frozen generator of diffusion model \cite{latentdiff}.

As for referenced NOFT\_2D, given a reference image $I_{Ref}$, we extract the image prompt using IP-Adapter \cite{ip} for consistent appearance transfer. Furthermore, we utilize the diffusion inversion method \cite{null} to recover the corresponding contextual latent of $I_{Ref}$. $NOFT_{\theta}^{2D}$ perturbs the inversed noise to generate diverse images:
\begin{equation}
\label{equ:2D_Ref}
I_{NOFT} =G_{\phi}^{2D*}(NOFT_{\theta}^{2D}(Inv(I_{Ref}), N_{Div}), I_{Ref})
\end{equation}
\subsubsection{NOFT\_3D}
TRELLIS \cite{trellis} compresses the 3D asset representation into a structured 3D latent similar to Latent Diffusion \cite{latentdiff}. It's possible for $NOFT_{\theta}^{3D}$ to implement the 3D tradeoff considering structural and textural preservation, along with the distribution diversity of 3D models and neural rendering \cite{nerf, 3dgs, 3dgs1}:
\begin{equation}
\label{equ:3D}
M_{NOFT} =G_{\phi}^{3D*}(NOFT_{\theta}^{3D}(N_{Oirg}, N_{Div}), \text{'S'}),
\end{equation}
where $G_{\phi}^{3D*}$ is the frozen generator of TRELLIS \cite{trellis}.

\subsection{Test-time noise finetune}
\label{otib}
We show the technical details of the noise information bottleneck along with Sinkhorn optimal transport of NOFT as follows:
\begin{equation}
\label{equ:2D_Ref}
N_{NOFT} =IB(N_{Orig}+\mathcal{F}_{SA}(N_{Orig}), N_{Div}),
\end{equation}
where $\mathcal{F}_{SA}$ is a Sinkhorn Attention module, as shown in Figure \ref{fig:pipeline}. 
\subsubsection{Noise information bottleneck}
\begin{figure}[htbp]
  \centering
  \makebox[\linewidth]{\includegraphics[width=\linewidth]{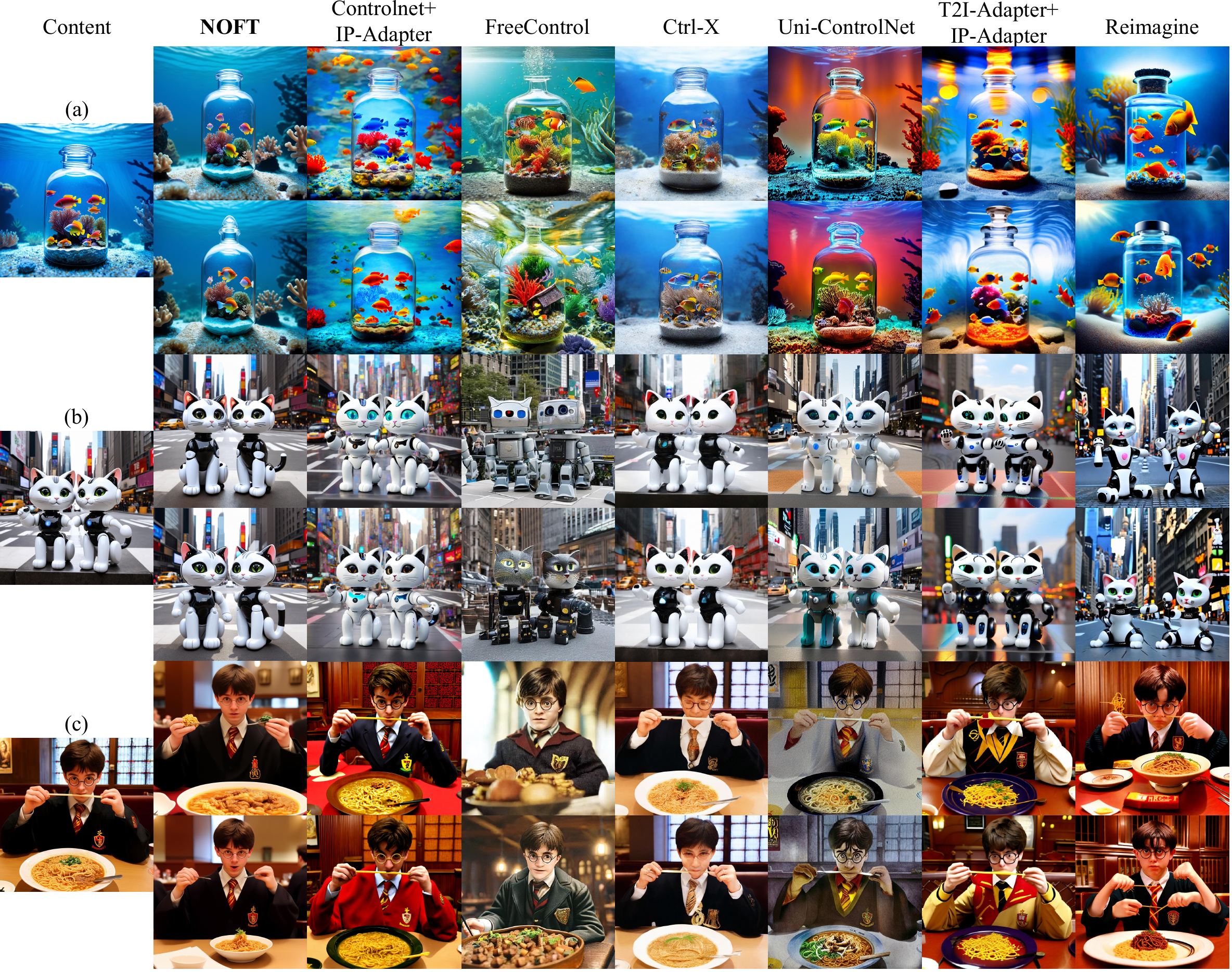}} 
  \vspace{-15.pt}
  \caption{Qualitative results of NOFT\_2D, ControlNet + IP Adapter \cite{controlnet, ip}, FreeControl \cite{freecontrol}, Ctrl-X \cite{ctrlx}, Uni-ControlNet \cite{uni}, T2I-Adapter + IP Adapter \cite{t2i-adapt, ip}, and Reimagine \cite{reimagine}. Zoom in for better observation. NOFT realizes more controllable image variations with high-fidelity content.}
  \label{fig:text}
  \vspace{-8.pt}
\end{figure}

As mentioned in Section \ref{Preliminaries}, implicit neural compression of information can be formulated as follows:
\begin{equation}
\label{equ:mib}
\mathop{\min}_{Z} \beta\mathbb{I}(R;Z),
\end{equation}
where $\mathbb{I}$ denotes the mutual information function, $Z$ is the manipulated feature derived from $R$. To realize high-fidelity content preservation and generation diversity, we adaptively learn a neural information filter $\lambda$. Given $R \sim \mathcal{N}(\mu_{G},\sigma_{G}^{2})$, 
 where $\mu_{G}$ and $\sigma_{G}$ represent the means and standard deviations of $R$. Then, the modulated manifold of 2D/3D asset can be formulated as follows \cite{iba}:
\begin{equation}
Z = \lambda R+(1-\lambda)\epsilon,
\label{equ:compress}
\end{equation}
where $Z$, $R$ and random Gaussian noise $\epsilon$ are from a consistent distribution $\mathcal{N}(\mu_{G},\sigma_{G}^{2})$ . The intent of NOFT is to improve representation diversity while implicitly adhering to the global content attributes of a certain scenario. If $\lambda$ is 0, the whole manifold will be replaced by $\epsilon$, which results in entire structure and appearance leakages. If $\lambda$ is 1, $Z$ excludes any form of diversity-inducing perturbations. Qualitative analyses are illustrated in Figure \ref{fig:text}, \ref{fig:ref}, and \ref{fig:abl_hard}.
\subsubsection{Sinkhorn Optimal Transport}
We impose a Sinkhorn Attention module $\mathcal{F}_{SA}$ in a spatial-OT view to improve contextual preservation of NOFT. First, we revisit the Optimal Transport that provides a mathematical framework for transporting probability distributions from the source to the target. Given discrete distributions as:
\begin{equation}
\mu = \sum_{i=1}^M \mu_i \delta_{x_i}, \quad \nu = \sum_{j=1}^N \nu_j \delta_{y_j}
\end{equation}
where $\mu, \nu$ are discrete probability measures, $\mu_i \geq 0$, $\nu_j \geq 0$ are probability masses ($\sum_i \mu_i = \sum_j \nu_j = 1$), $\delta_x$ denotes the Dirac delta function centered at point $x$, $M$ and $N$ are the number of support points. The original OT problem finds a transport plan $\mathbf{T}^*$ that minimizes the total transportation cost, which is computationally intensive. The Sinkhorn algorithm \cite{sinkhorn, otseg} equips OT with an entropy regularization term:
\begin{equation}
\mathbf{T}^* = \arg\min_{\mathbf{T} \in \Pi(\mu,\nu)} \langle \mathbf{T}, \mathbf{C} \rangle_F - \epsilon H(\mathbf{T}),
\end{equation}
where $\mathbf{T} \in \mathbb{R}^{M \times N}$ is the transport matrix with $\mathbf{T}_{ij}$ specifying how much mass moves from $x_i$ to $y_j$, $\mathbf{C} \in \mathbb{R}^{M \times N}$ is the cost matrix where $\mathbf{C}_{ij} = d(x_i,y_j)$, $\Pi(\mu,\nu) = \{\mathbf{T} \geq 0 \mid \mathbf{T}\mathbf{1^{N}} = \mu, \mathbf{T}^\top\mathbf{1^{M}} = \nu\}$ defines the set of admissible transport plans, $\langle \cdot, \cdot \rangle_F$ denotes the Frobenius inner product. Moreover, $\epsilon > 0$ is the regularization strength, $H(\mathbf{T}) = -\sum_{ij} \mathbf{T}_{ij}\log \mathbf{T}_{ij}$ is the entropy of the transport plan.
The Sinkhorn algorithm solves this through iterative Bregman projections:
\begin{algorithm}[H]
\caption{Classical Sinkhorn Iteration}
\begin{algorithmic}[1]
\State Initialize $\mathbf{K} = \exp(-\mathbf{C}/\epsilon)$ \Comment{Gibbs kernel}
\Repeat
    \State $\mathbf{u} \gets \mu \oslash (\mathbf{K}\mathbf{v})$ \Comment{Row scaling ($\oslash$: element-wise division)}
    \State $\mathbf{v} \gets \nu \oslash (\mathbf{K}^\top\mathbf{u})$ \Comment{Column scaling}
\Until{Convergence}
\State \Return $\text{diag}(\mathbf{u})\mathbf{K}\text{diag}(\mathbf{v})$ \Comment{Optimal transport plan}
\end{algorithmic}
\end{algorithm}
\vspace{-15pt}
where $\mathbf{u} \in \mathbb{R}^M$, $\mathbf{v} \in \mathbb{R}^N$ are scaling vectors. Convergence typically measured by $\|\mathbf{T}\mathbf{1^{N}} - \mu\|_1 < \text{tol}$. In our NOFT algorithm, the Sinkhorn Attention module is as follows:
\begin{algorithm}[H]
\caption{Sinkhorn-Attention Forward Pass}
\begin{algorithmic}[1]
\State \textbf{Input:} Feature map $X \in \mathbb{R}^{B \times C \times H \times W}$
\State $Q = \text{Conv\_Nd}(X)$, $K = \text{Conv\_Nd}(X)$, $V = \text{Conv\_Nd}(X)$ \Comment{Learnable projections}
\State $A = QK^\top/\sqrt{C}$ \Comment{Attention logits}
\For{$k=1$ to $n_{iters}$}
    \State $A = A - \text{LogSumExp}(A, \text{dim}=2)$ \Comment{Row normalization}
    \State $A = A - \text{LogSumExp}(A, \text{dim}=1)$ \Comment{Column normalization}
\EndFor
\State $\mathbf{T} = \exp(A)$ \Comment{Optimal attention weights}
\State \Return $\mathbf{T}V$ \Comment{Transport applied to values}
\end{algorithmic}
\end{algorithm}
\vspace{-15pt}
where $Q,K,V \in \mathbb{R}^{B \times (HW) \times C}$ are Query, Key, Value tensors, respectively. $A \in \mathbb{R}^{B \times (HW) \times (HW)}$ is Attention logits matrix, $\text{LogSumExp}(A)_i = \log\sum_j \exp(A_{ij})$, and $\mathbf{T}$ is Doubly-stochastic attention matrix. Our transport solution is established through:
\begin{equation}
\mathbf{T}_{ij} = \exp(\underbrace{\frac{q_i^\top k_j}{\sqrt{C}}}_{\text{Transport cost}} - \underbrace{\alpha_i - \beta_j}_{\text{Sinkhorn scalars}})
\end{equation}
where $\alpha$ and $\beta$ are row and column normalization factors, respectively. The division by $\sqrt{C}$ stabilizes gradient flow.

\subsection{Training loss}
\label{loss}
Training losses contain pixel-level reconstruction loss and manifold-level information compression loss. As for noise consistency loss, the pixel-level supervision for $N_{NOFT}$ is formulated as MSE loss that demonstrates a powerful content preservation function \cite{diffusion, latentdiff, word, dreambooth, hyperdreambooth}:
\begin{equation}
\begin{aligned}
\mathcal{L}_{noise} = ||N_{NOFT}-X_{Orig}||_{2}^{2}.
\end{aligned}
\end{equation}
For Gaussian distribution $\mathcal{N}(\mu, \sigma^{2})$ and $\mathcal{N}(0,1)$, KL divergence is formulated as:
\begin{equation}
\begin{aligned}
\mathcal{D}_{KL}[N(\mu, \sigma^{2})\Vert N(0,1)]
                   = -\frac{1}{2}[log(\sigma)^{2}-(\sigma)^{2}-(\mu)^{2}+1].
\end{aligned}
\end{equation}

\begin{figure}[t]
  \centering
  \makebox[\linewidth]{\includegraphics[width=\linewidth]{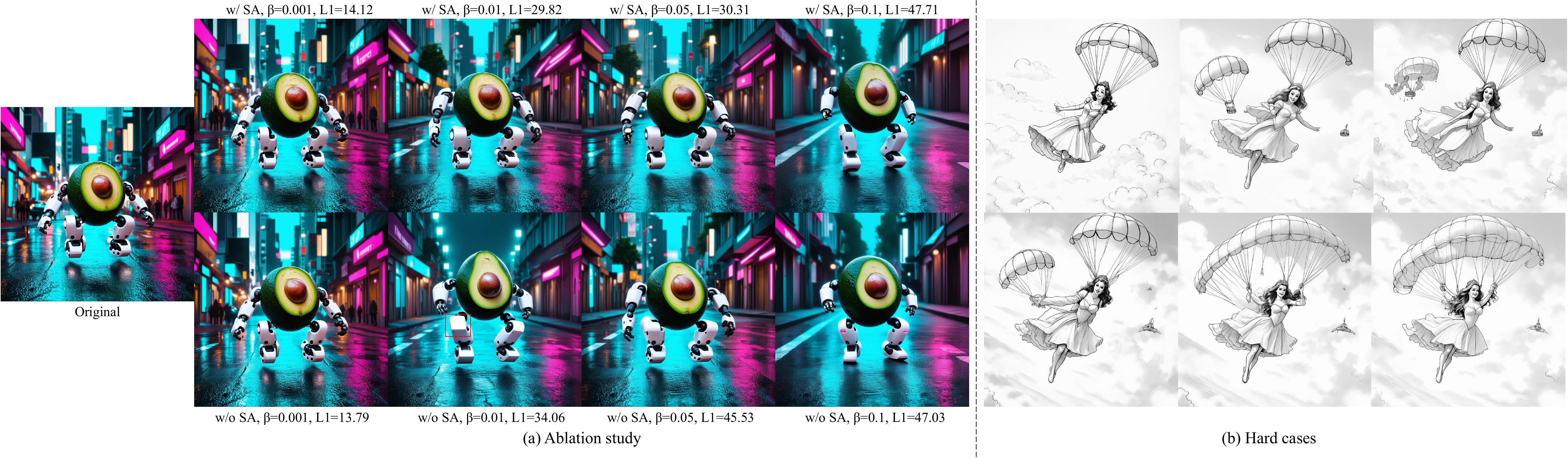}} 
  \vspace{-15.pt}
  \caption{(a) NOFT variants show that methods w/ SA preserve better appearance statistics than those w/o SA. Higher $\beta$ usually intentionally relaxes contextual constraints but boosts the diversity (Figure \ref{fig:self_com}). Zoom in for better observation. (b) There are some artifacts for sketch images, while the body pose of the princess is maintained with diverse head poses.}
  \label{fig:abl_hard}
  \vspace{-15.pt}
\end{figure}

Our framework eliminates the need for feature mean/variance pre-calculation by leveraging the predefined properties of Gaussian noise ($\mu_{G}$=0, $\sigma_{G}$=1). As for our case mentioned in Equ. \ref{equ:3}, the distribution of $p(Z|R)$ is accessed as $\mathcal{N}[\lambda R,(1-\lambda)^{2}]$ according to Equ. \ref{equ:compress}. We normalize $p(Z|R)$ along with $q(Z)$ using $\mu_{G}$ and $\sigma_{G}$, then the information compression metric of NOFT is:
\begin{equation}
\mathcal{L}_{info} = \mathbb{I}(Z;R)= KL[p(Z|R)\Vert q(Z)]= -\frac{1}{2}[log(1-\lambda)^{2}-(1-\lambda)^{2}-(\lambda R)^{2}+1],
\end{equation}
Finally, the total loss of NOFT is formulated as:
\begin{equation}
\mathcal{L}_{\emph{NOFT}}= \beta\mathcal{L}_{info}+  \mathcal{L}_{noise},
\label{con:20}
\end{equation}
where $\beta$ is the content-diversity tradeoff weight (Fig. \ref{fig:abl_hard}).

\vspace{-5.pt}
\begin{figure}[htbp]
  \centering
  \makebox[\linewidth]{\includegraphics[width=\linewidth]{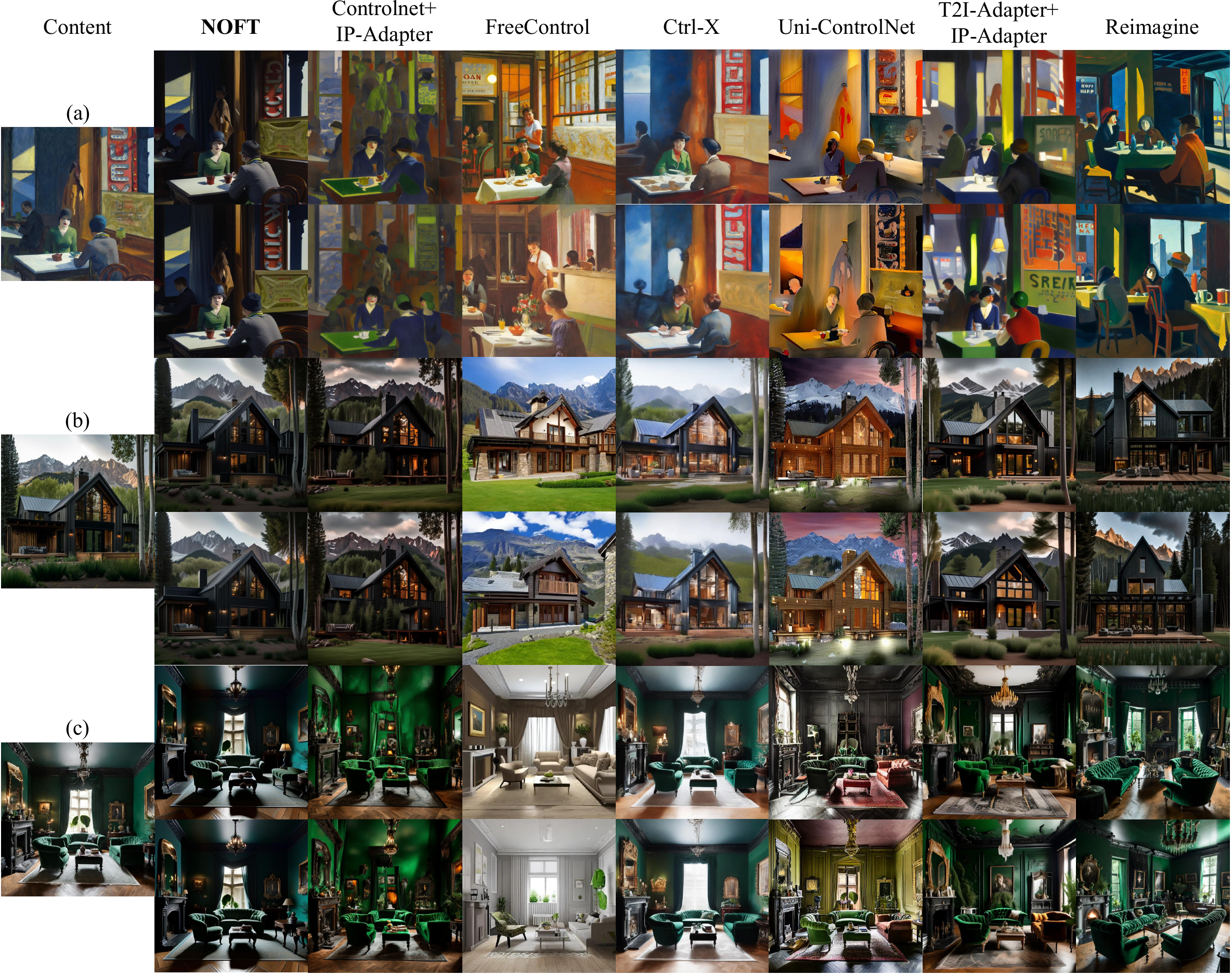}} 
  \vspace{-15.pt}
  \caption{Qualitative results of NOFT\_2D\_Ref, ControlNet \cite{controlnet, ip}, FreeControl \cite{freecontrol}, Ctrl-X \cite{ctrlx}, Uni-ControlNet \cite{uni}, T2I-Adapter \cite{t2i-adapt, ip} and Reimagine \cite{reimagine} on datasets \cite{ctrlx}. Previous methods generate diverse images based on structure and texture signals from the same source.}
  \label{fig:ref}
  \vspace{-12.pt}
\end{figure}
\section{Experiments}
\label{others}
Through comprehensive qualitative and quantitative evaluations, we validate NOFT's dual capability in maintaining content fidelity while enhancing generation diversity for digital asset creation. Additional results are provided in Appendix A.
\paragraph{Training Protocol.} We train our NOFT on Gaussian noise tensors with corresponding dimension shape of different architectures, e.g., $4*64*64$ \cite{latentdiff}, $16*128*128$ \cite{sd3}, $8*16*16*16$ \cite{trellis}. $N_{Orig}$ and $N_{Div}$ are random noises in each training step. As for NOFT\_3D, we utilize 3D convolutions for SA and IB modules. We train NOFT for 20k iterations with one NVIDIA RTX 4090 GPU. The training batch size is set to 1. During training, we employ Adam \cite{adam} with $2*10^{-3}$ learning rate. We set $\beta=0.01$ for mild diversity (a,b in Figure \ref{fig:text}), $\beta=0.1$ for substantial diversity (Figure \ref{fig:self_com}, c in Figure \ref{fig:text}), and $\beta=1$ for diversity with reference constraints (Figrue \ref{fig:ref}). 
\paragraph{Baselines.}
There are several state-of-the-art controllable synthesis methods based on diffusion models. ControlNet \cite{controlnet} and T2I-Adapter \cite{t2i-adapt} align diffusion priors to the external control structures. We further apply IP-Adapter \cite{ip} to them for better textural transfer. These methods present low topological flexibility with restriction by the explicit structure alignment, and limited textural fidelity with global appearance control. FreeControl \cite{freecontrol} has large-scale content variance due to imprecise structure and appearance representations (col 4 in Figure \ref{fig:text}\& \ref{fig:ref}). Ctrl-X \cite{ctrlx} provides too-strict structure and appearance alignments, and there are texture distortions. Uni-ControlNet \cite{uni} also suffers from the global appearance representation (col 6 in Figure \ref{fig:text}\& \ref{fig:ref}). Stable diffusion Reimagine \cite{reimagine} produces uncontrollable content layout, despite high image quality and diversity (col 8 in Figure \ref{fig:text}\& \ref{fig:ref}). We evaluate all methods on SDXL v1.0 \cite{sdxl} when workable and on their pre-configured base models otherwise. 

\paragraph{Evaluation metrics.}
Tab. \ref{com} shows a quantitative comparison of natural images of datasets \cite{ctrlx}. The objective metrics include DINO ViT self-similarity \cite{splice}, DINO-I \cite{dreambooth}, and pixel-wise L1 distance between the source image and generated image. L1 attempts to measure both contextual preservation and detail diversity. Note that NOFT shows consistent superiority on self-sim and DINO-I. Meanwhile, the subjective metrics consist of quality, fidelity, and diversity subject to fidelity. NOFT achieves comparable user preference.
\paragraph{Qualitative results.}
NOFT only learn noise representation supervised by itself based on OTIB. Visually comparable results demonstrate that our implicit NOFT is a better workbench for highly correlated asset editing. As shown in Figure \ref{fig:self_com} and (c) of Figure \ref{fig:text}, NOFT implicitly changes the size, position, and local semantics of objects, e.g., 'cat', 'cheese', 'beef noodle bowl'. More results are shown in the Appendix.
\begin{table}
  \caption{NOFT outperforms other SOTA methods in structure and appearance alignments, measured by DINO ViT self-similarity \cite{splice} and DINO-I \cite{dreambooth}. We report the inference time of NOFT\_2D and NOFT\_2D\_Ref where diffusion inversion \cite{null} is time-consuming. Moreover, NOFT exhibits competitive human preference percentages.}
  \label{com}
  \centering
  \scalebox{0.65}{
  \begin{tabular}{lcc|ccc|cccc}
    \toprule
    Methods     & Training     & Inference time (s)  & self-sim $\downarrow$ & DINO-I $\uparrow$ & L1 & Quality $\uparrow$ &Fidelity $\uparrow$&Diversity (\small s.t. \small Fidelity)$\uparrow$\\
    \midrule
    Uni-ControlNet \cite{uni}& \usym{2713}  &  10.6  & 0.045& 0.555 &56.41 &80\% & 72\% &78\%\\
    ControlNet + IP Adapter \cite{controlnet, ip}  & \usym{2713} &  8.1 & 0.068&  0.656& 46.06 &50\% &63\% &79\%\\
    T2I-Adapter + IP Adapter \cite{t2i-adapt, ip} &   \usym{2713}  &4.2 &0.055 &  0.603& 50.45&71\% &60\% &76\%\\
    Ctrl-X \cite{ctrlx}  & \usym{2715} & 14.9  & 0.057& 0.686 & 37.07  &85\% & \textbf{93\%}         &72\%                   \\
    FreeControl \cite{freecontrol} & \usym{2715} & 21.5  & 0.058&0.572  &  85.45        &68\% &54\%    &64\%       \\
    Reimagine \cite{reimagine}   & \usym{2713} &  10.1 & 0.073& 0.753 & 64.12 &\textbf{93\%} &34\%         &48\%            \\
    \textbf{NOFT (ours)}  & \usym{2713} &  7.3 / 27.2 &\textbf{0.038} & \textbf{0.841} &  41.58   & 90\% &90\%       &\textbf{92\%}       \\
    \bottomrule
  \end{tabular}}
  \vspace{-5pt}
\end{table}
\begin{figure}[t]
  \centering
  \makebox[\linewidth]{\includegraphics[width=\linewidth]{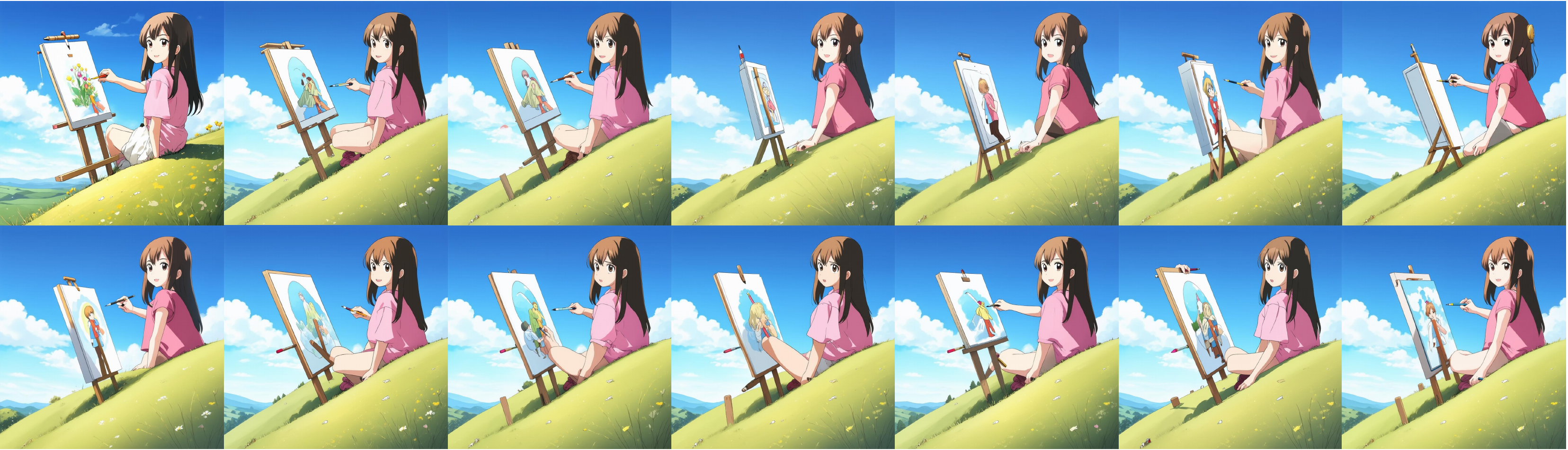}} 
  \vspace{-15.pt}
  \caption{NOFT helps diffusion model to realize content-diversity tradeoff where the girl exhibits different facial expressions and hand poses, and the drawing boards display artworks with both high diversity and perceptual coherence. The left-top is the source image. Zoom in for better observation.}
  \label{fig:diverse}
  \vspace{-15.pt}
\end{figure}
\paragraph{Ablation Study}
As shown in Figure \ref{fig:abl_hard} (a), the NOFT variants without Sinkhorn Attention fail to capture local structure and appearance patterns (red boxes in col 3\&4). The context-diversity tradeoff weight $\beta$ controls the structure and appearance leakage in an adaptive way. 
\paragraph{Limitations}
There are some hard cases, such as sparse sketch images in Figure \ref{fig:abl_hard} (b). There are some artifacts for the local structures of small objects, e.g., hands, and the people in the far distance.
\vspace{-8.pt}
\section{Conclusion}
Our proposed noise finetune (NOFT) completely disentangles highly correlated concept representation learning from both dependencies of training asset data or external control signals, and the pre-trained T2I model finetune. We present an efficient and effective OTIB module that provides a trade-off of preservation of topology and texture, as well as semantic diversity. Compared with state-of-the-art structure-aligned and appearance-aligned approaches, comprehensive experimental analyses demonstrate that NOFT is promising to be the first effective plug-and-play implicit controller for pre-trained T2I models with remarkable context consistency and content diversity.
\paragraph{Broader impacts.} Our method provides a robust editor for both images and 3D models. While its primary advantage lies in assisting designers, animators, and 3D modelers in asset creation, the potential for malicious manipulation of visual assets necessitates mandatory watermarking in practical applications.

\bibliographystyle{IEEEtran}
\small
\bibliography{bi}

\newpage
\appendix

\section{Additional results}
\begin{figure}[htbp]
  \centering
  \makebox[\linewidth]{\includegraphics[width=1\linewidth]{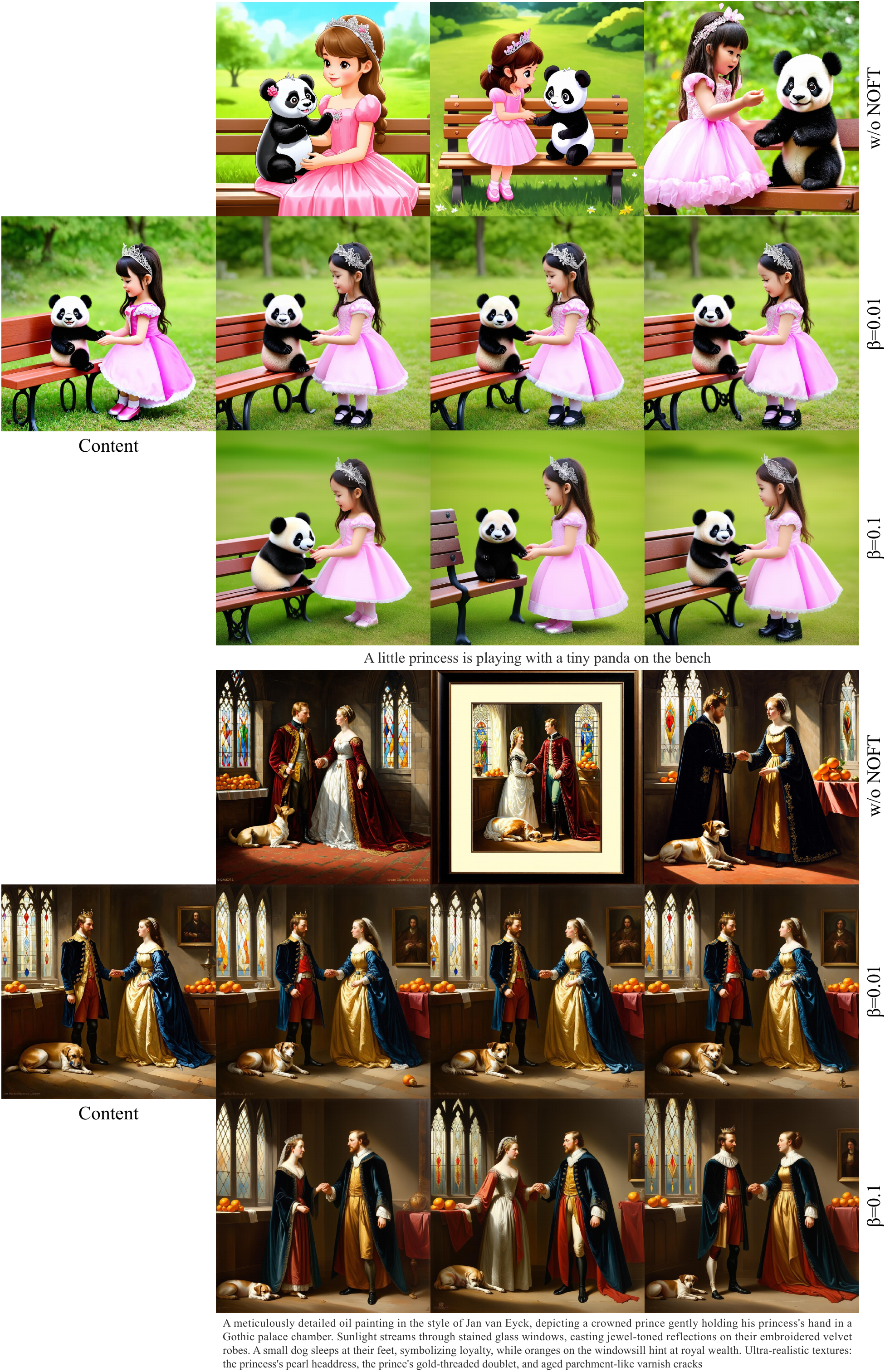}} 
  \caption{NOFT effectively controls the structure and appearance of the content. Smaller tradeoff weight $\beta$ puts content on a slight adjustment workbench, while larger $\beta$ changes the content more obviously, but maintains the scene layout. }
  \label{fig:role}
\end{figure}

\begin{figure}[htbp]
  \centering
  \makebox[\linewidth]{\includegraphics[width=1.3\linewidth]{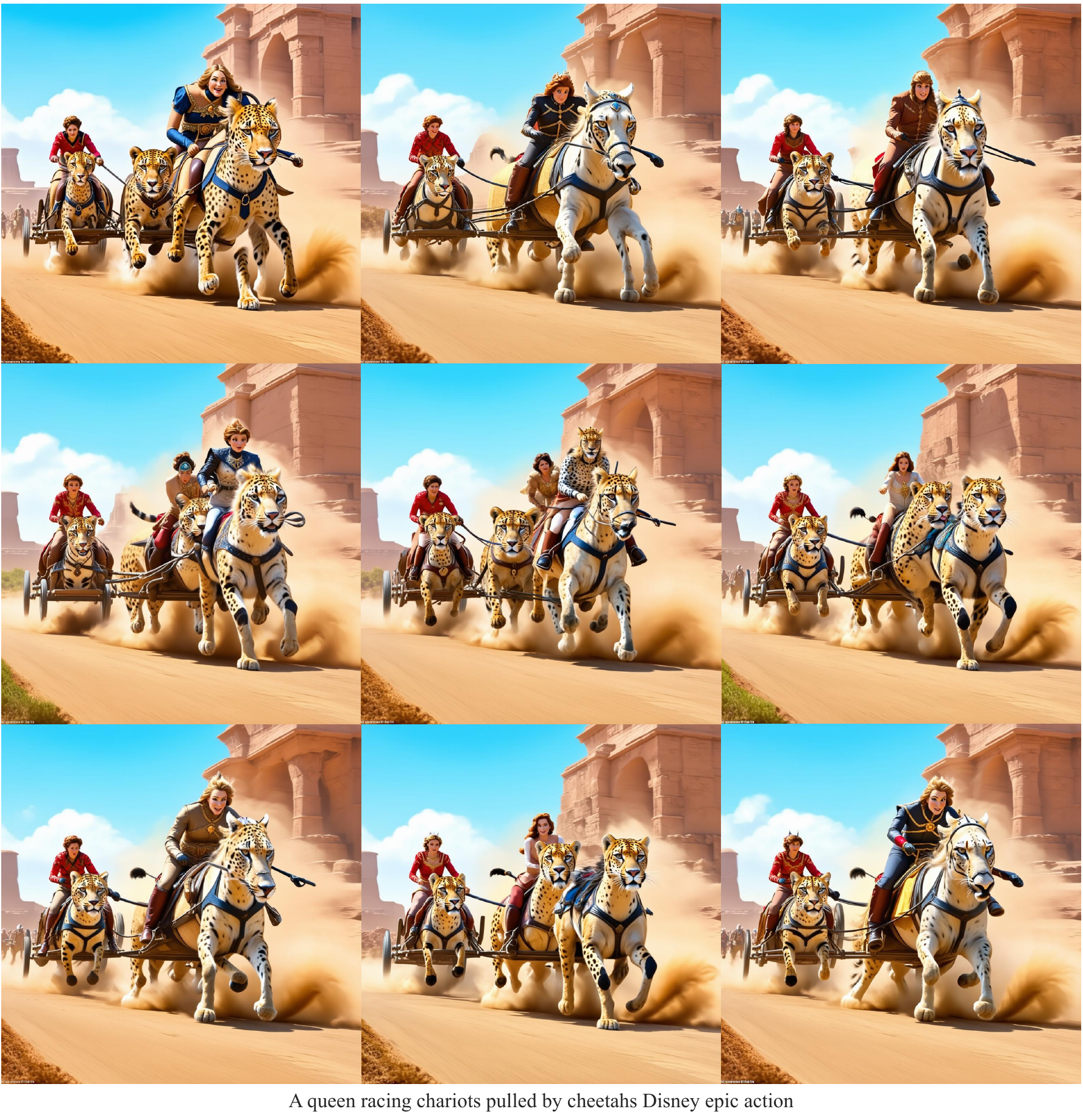}} 
  \caption{Substantial diversity visualization where the queen and cheetahs have various structures and appearances in different generated images based on NOFT.}
  \label{fig:su4}
\end{figure}

\begin{figure}[htbp]
  \centering
  \makebox[\linewidth]{\includegraphics[width=\linewidth]{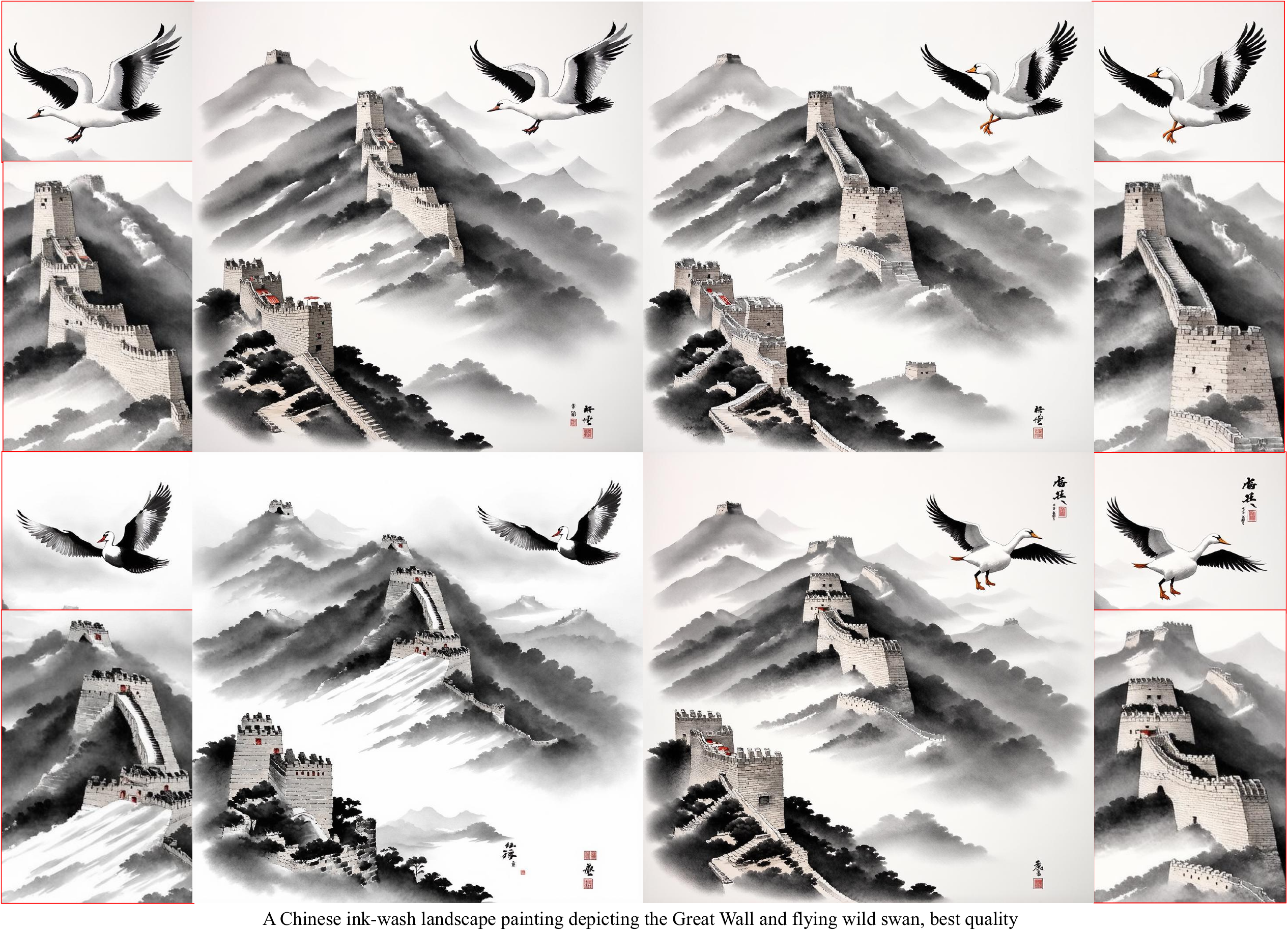}} 
  \caption{Image variants of the teaser figure \ref{fig:teaser} under magnified observation.}
  \label{fig:zoom}
\end{figure}

\begin{figure}[htbp]
  \centering
  \makebox[\linewidth]{\includegraphics[width=0.7\linewidth]{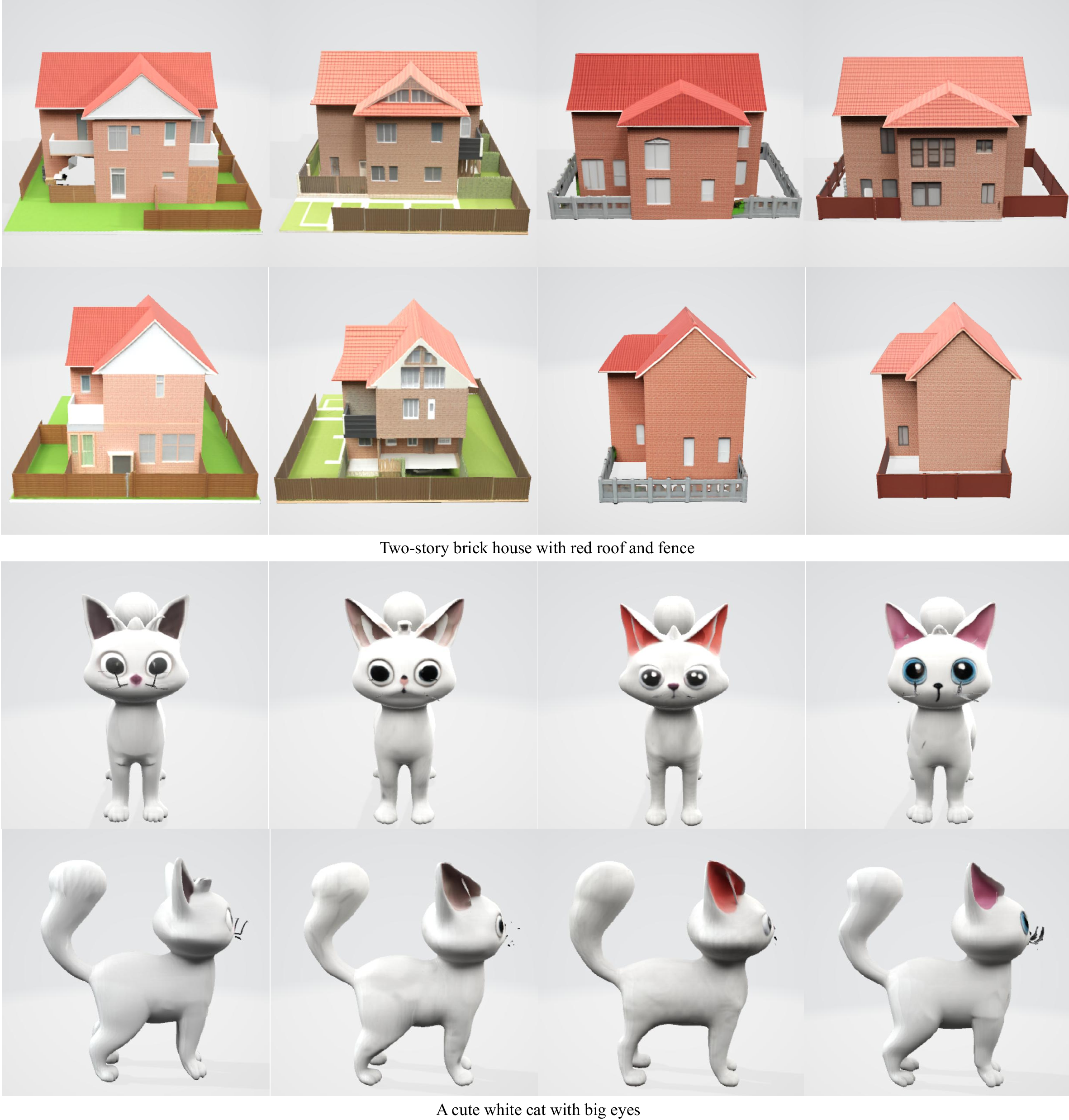}} 
  \caption{More qualitative results of NOFT\_3D based on TRELLIS \cite{trellis}.}
  \label{fig:3d_sup}
\end{figure}

\begin{figure}[htbp]
  \centering
  \makebox[\linewidth]{\includegraphics[width=1.3\linewidth]{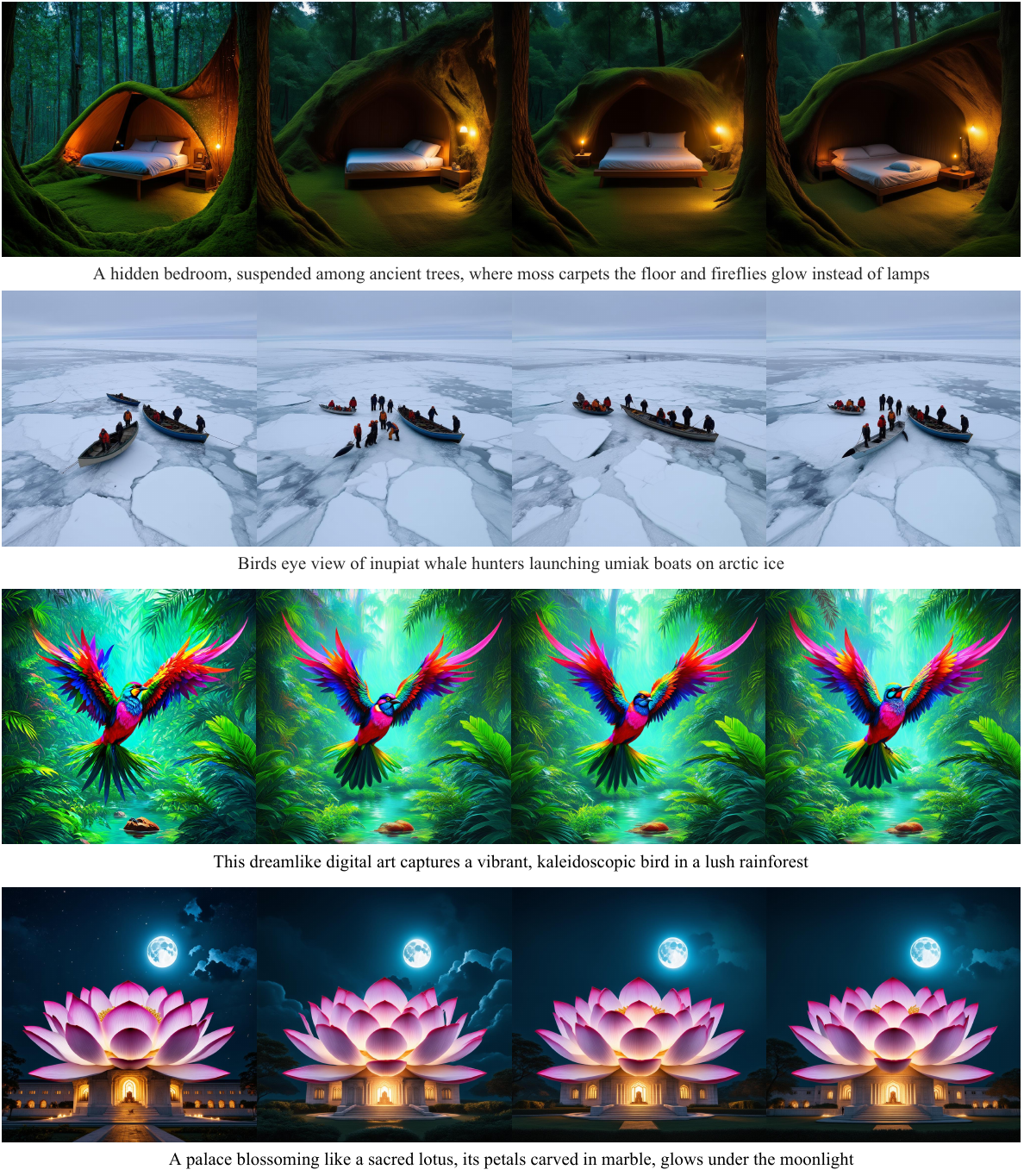}} 
  \caption{Additional visual results of NOFT\_2D based on SD3 \cite{sd3}.}
  \label{fig:su1}
\end{figure}

\begin{figure}[htbp]
  \centering
  \makebox[\linewidth]{\includegraphics[width=1.3\linewidth]{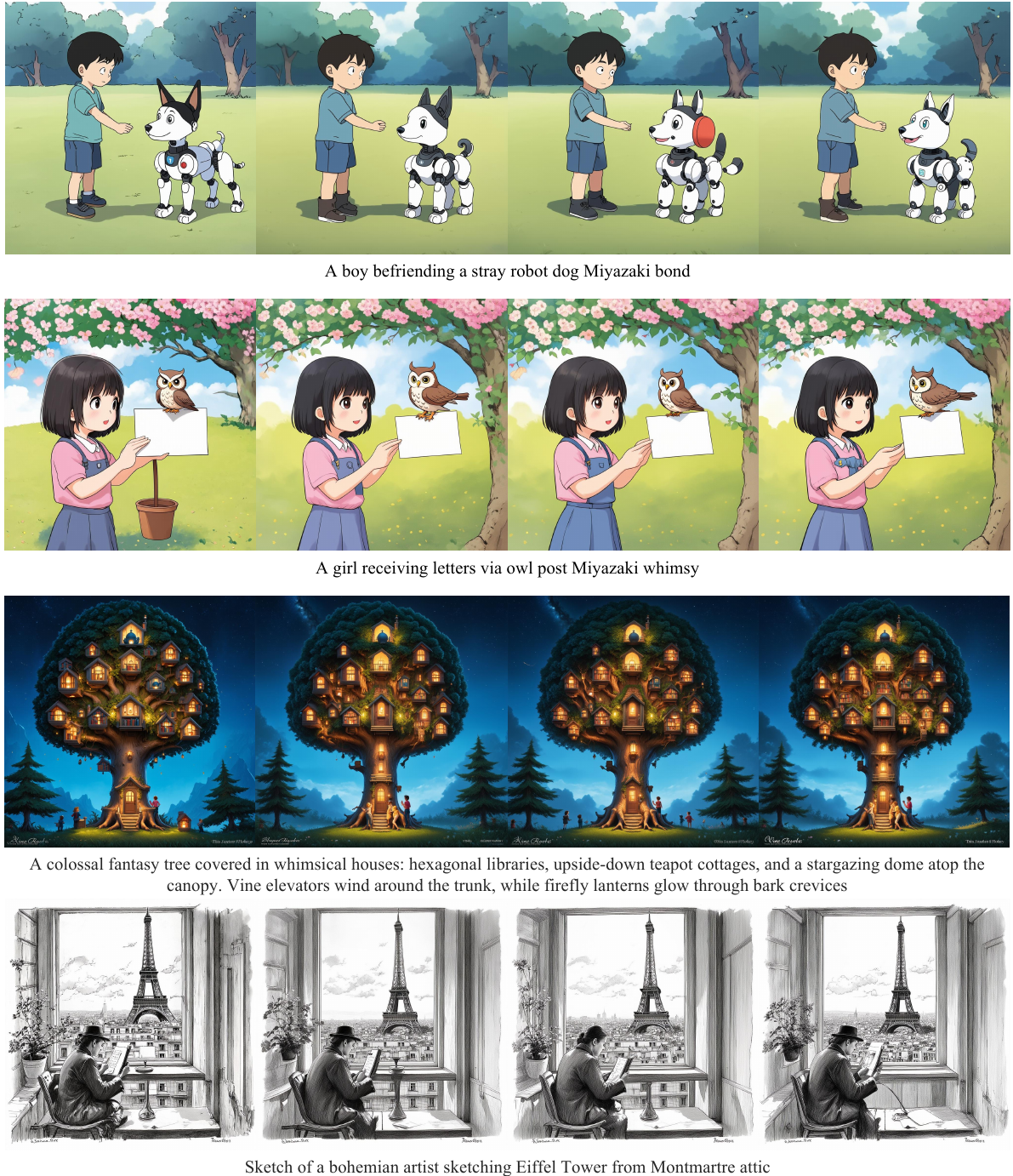}} 
  \caption{Additional visual results of NOFT\_2D based on SD3 \cite{sd3}.}
  \label{fig:su2}
\end{figure}

\begin{figure}[htbp]
  \centering
  \makebox[\linewidth]{\includegraphics[width=1.1\linewidth]{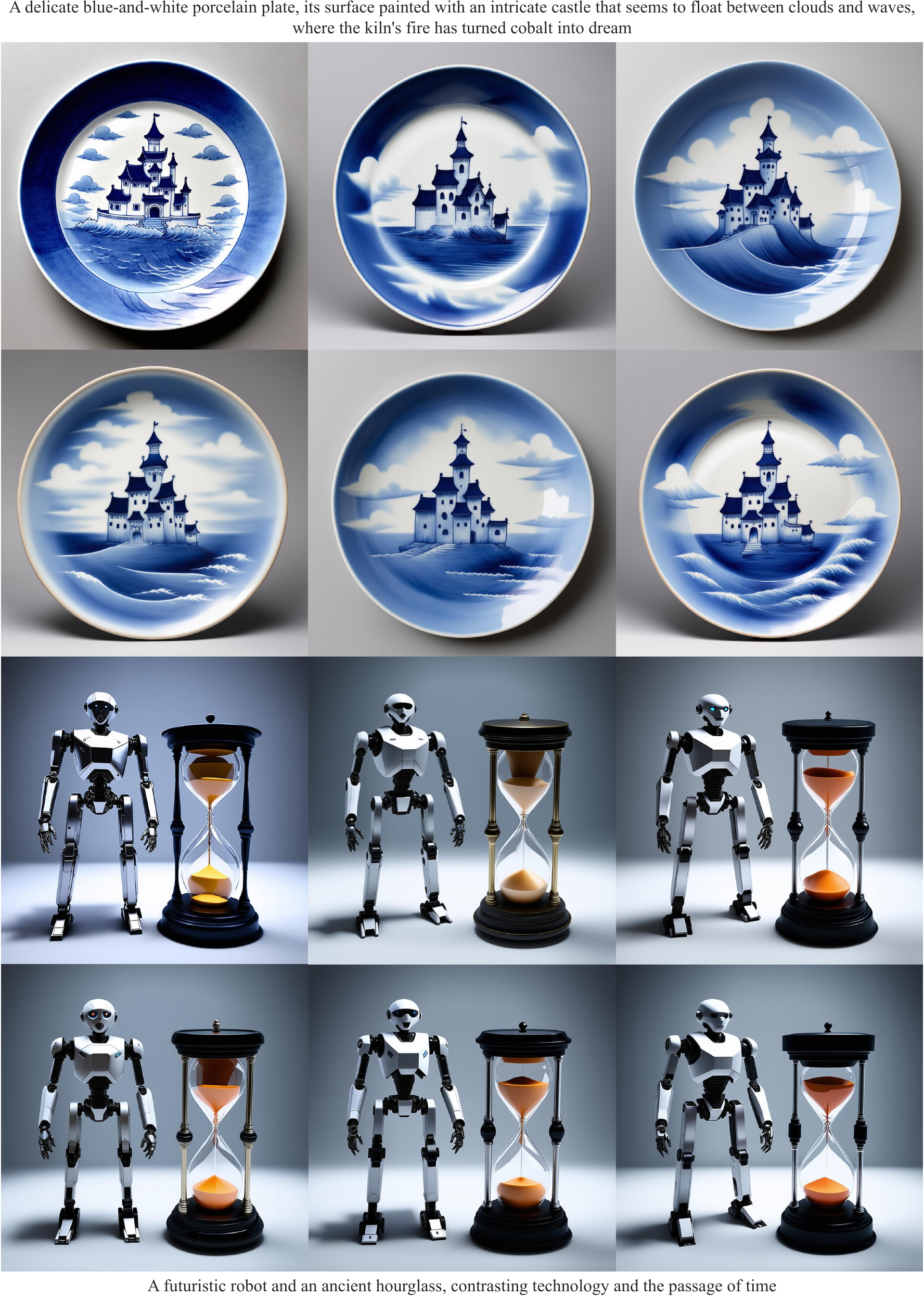}} 
  \caption{Additional visual results of NOFT\_2D based on SD3 \cite{sd3}.}
  \label{fig:su3}
\end{figure}

\begin{figure}[htbp]
  \centering
  \makebox[\linewidth]{\includegraphics[width=1.3\linewidth]{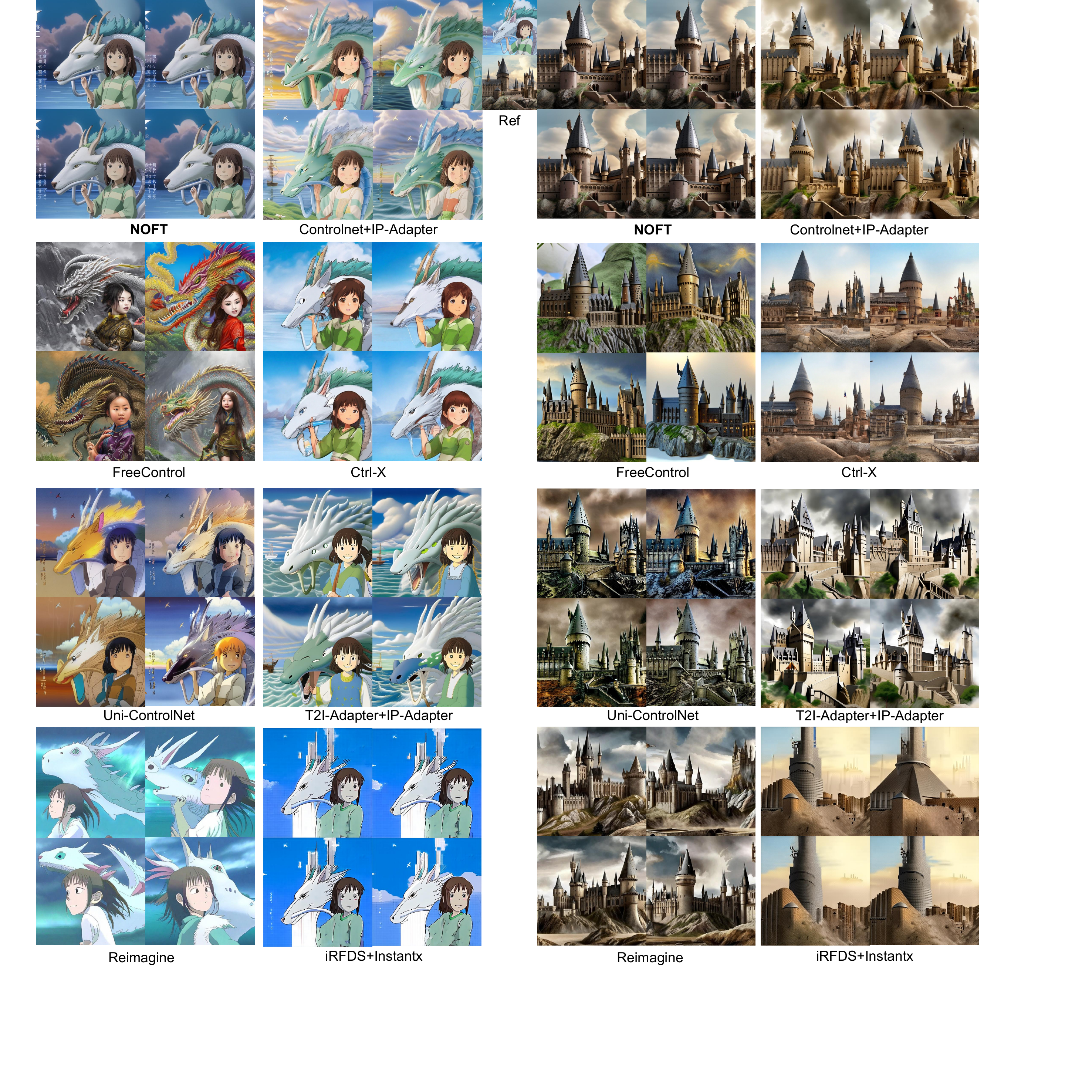}} 
  \caption{Qualitative results of NOFT\_2D\_Ref, ControlNet \cite{controlnet, ip}, FreeControl \cite{freecontrol}, Ctrl-X \cite{ctrlx}, Uni-ControlNet \cite{uni}, T2I-Adapter \cite{t2i-adapt, ip}, Reimagine \cite{reimagine} and iRFDS \cite{flowprior} on the wild images.}
  \label{fig:qx}
\end{figure}

\begin{figure}[htbp]
  \centering
  \makebox[\linewidth]{\includegraphics[width=1.3\linewidth]{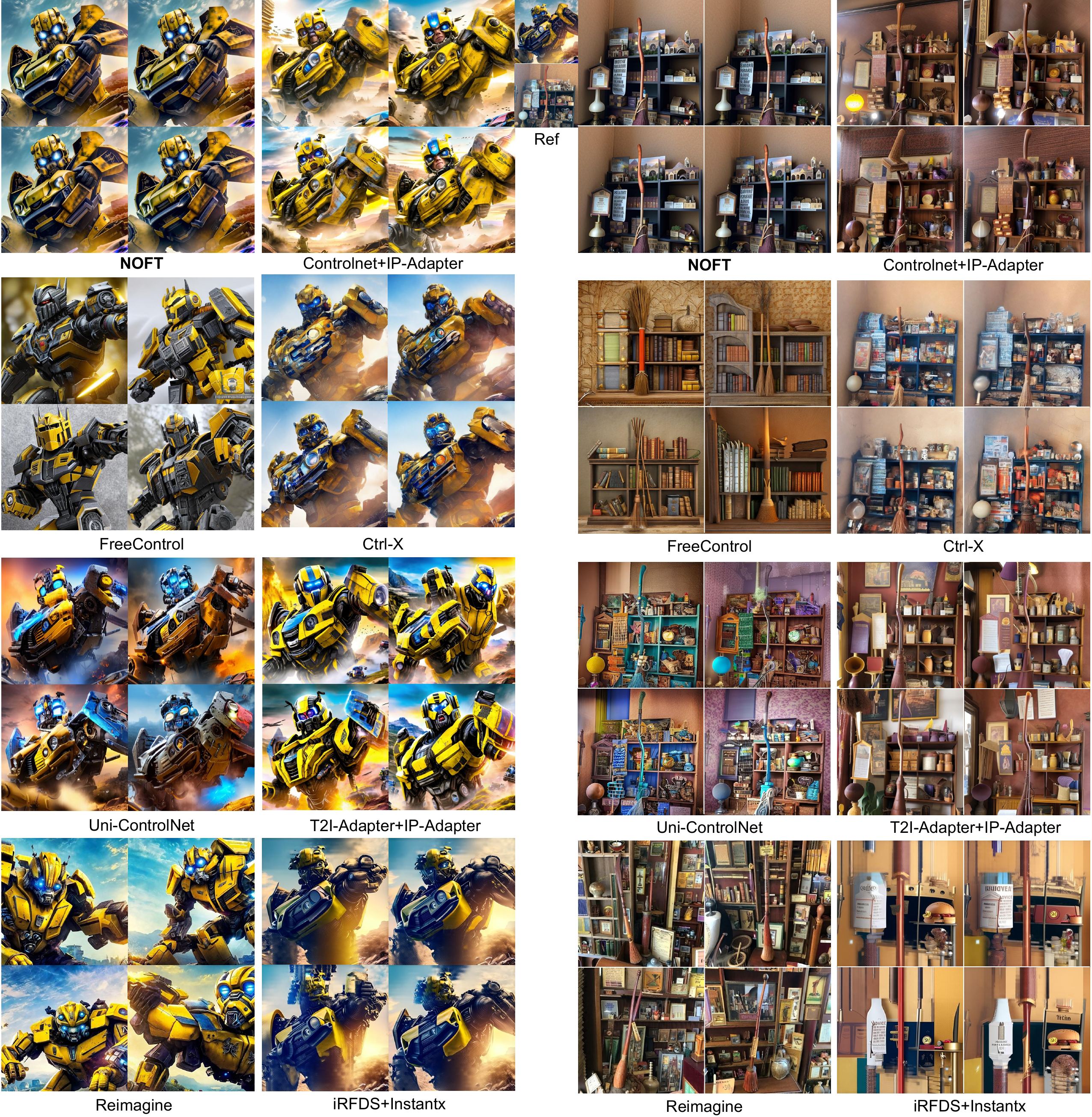}} 
  \caption{Qualitative results of NOFT\_2D\_Ref, ControlNet \cite{controlnet, ip}, FreeControl \cite{freecontrol}, Ctrl-X \cite{ctrlx}, Uni-ControlNet \cite{uni}, T2I-Adapter \cite{t2i-adapt, ip}, Reimagine \cite{reimagine} and iRFDS \cite{flowprior} on the wild images.}
  \label{fig:qtz}
\end{figure}

In this section, we provide additional qualitative results of 2D (Figure \ref{fig:su4}, \ref{fig:zoom}, \ref{fig:su1}, \ref{fig:su2}, \ref{fig:su3}, \ref{fig:qx}, \ref{fig:qtz}) or 3D asset (Figure \ref{fig:3d_sup}) creation based on NOFT. Figure \ref{fig:role} indicates the workable function of OTIB to conduct controllable diversity implicitly. Note that the detailed differences for small $\beta$ are not obvious. Please zoom in sufficiently and observe patiently.
\paragraph{Model select} As for NOFT\_2D\_Ref, we use Realistic\_Vision\_\ V4.0\_noVAE for diffusion inversion and denoising, with ip-adapter-plus\_sd15 for appearance transfer. The VAE module is from stabilityai-stable-diffusion-2-1-base. In Figure \ref{fig:qx} and \ref{fig:qtz}, iRFDS+Instantx uses the checkpoint of InstantX-SD3.5-Large-IP-Adapter. In Figure \ref{fig:teaser}, images of NOFT\_2D are synthesized based on the checkpoint of Stable Diffusion v2-1\_512-ema-pruned. 

\begin{figure}[htbp]
  \centering
  \makebox[\linewidth]{\includegraphics[width=0.85\linewidth]{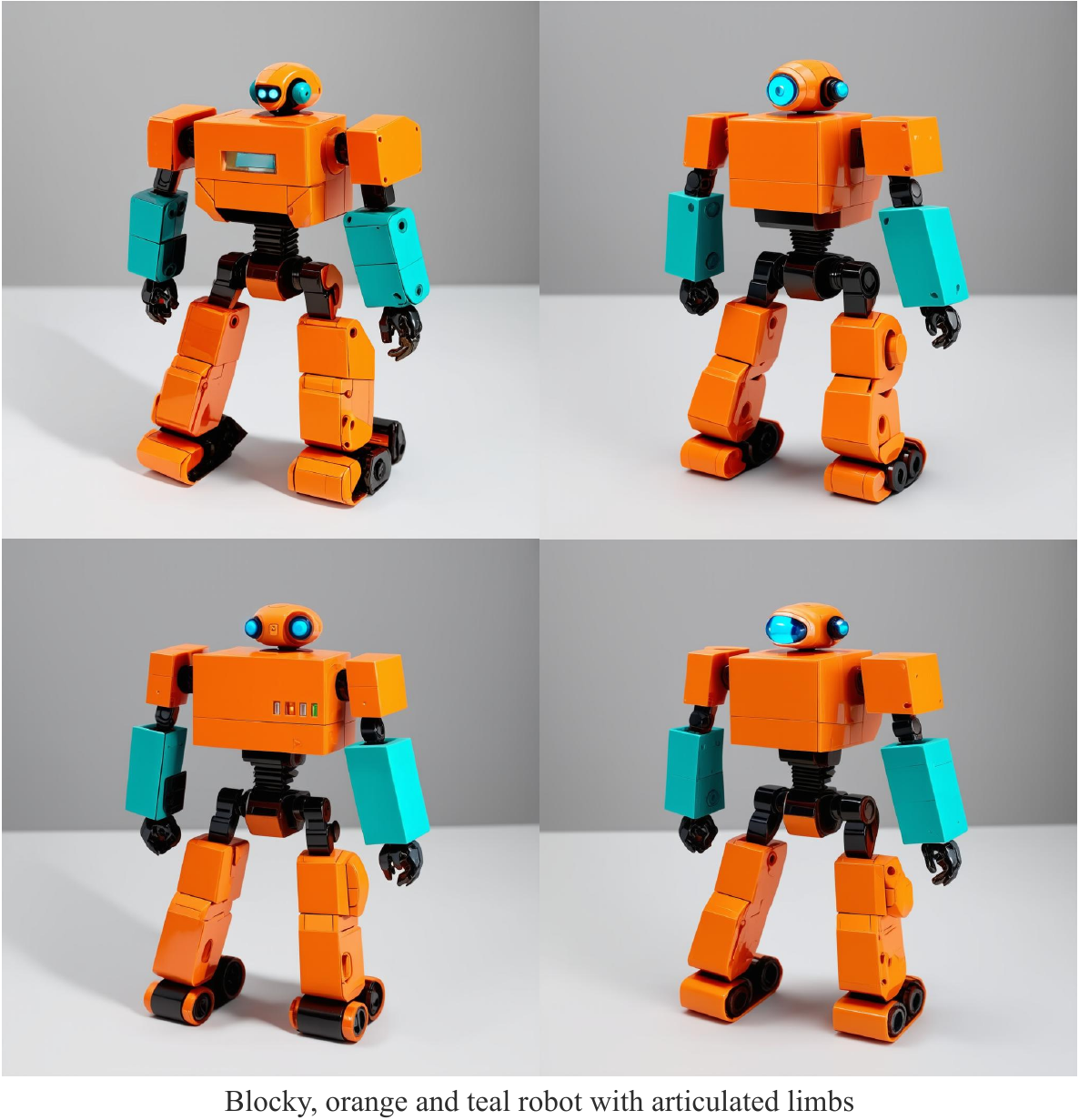}} 
  \caption{The first stage based on NOFT\_2D of the NOFT\_3D\_Img in Figure \ref{fig:teaser}.}
  \label{fig:sup_robot}
\end{figure}

\begin{figure}[htbp]
  \centering
  \makebox[\linewidth]{\includegraphics[width=1.3\linewidth]{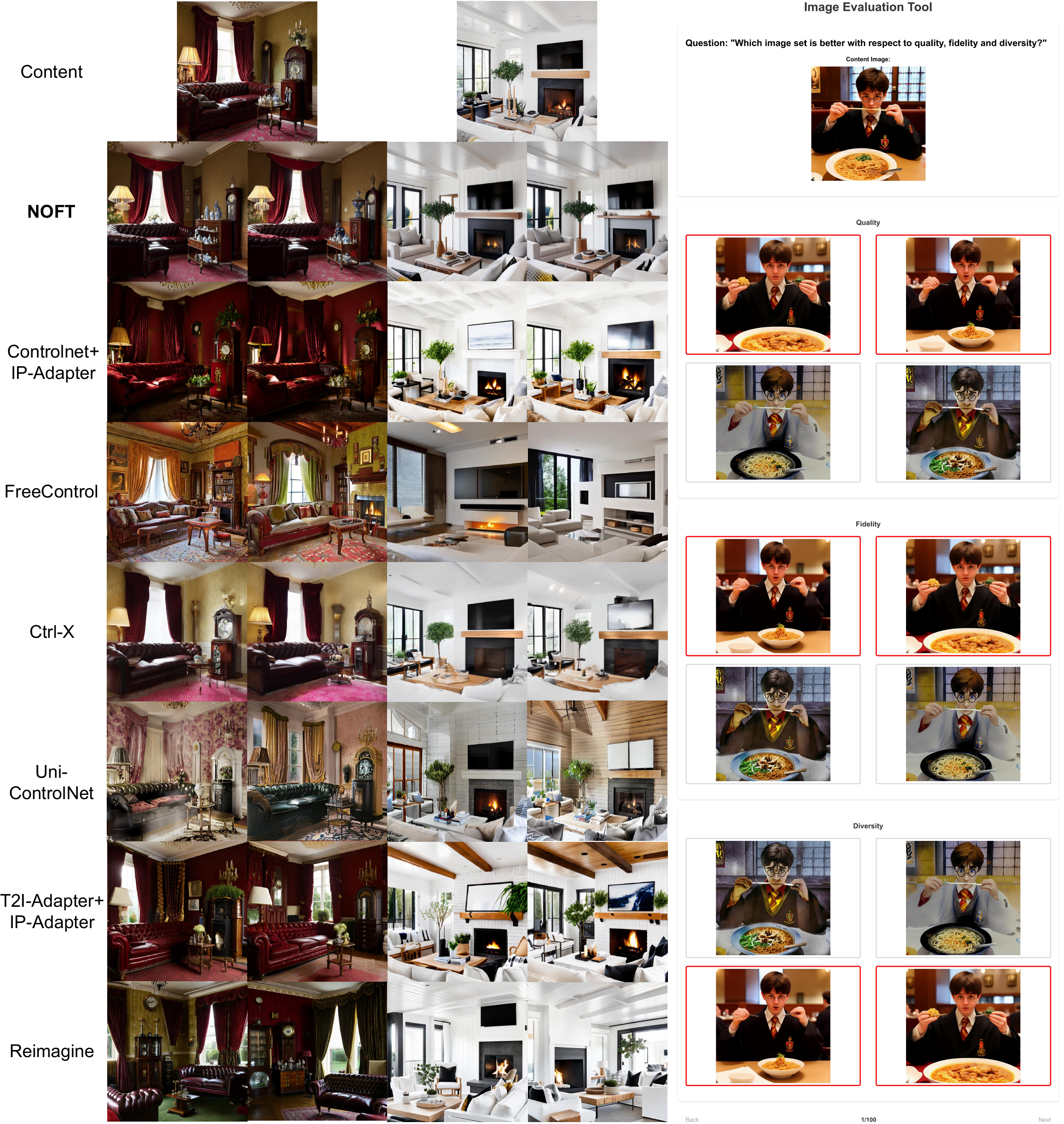}} 
  \caption{(a) Additional qualitative results of NOFT\_2D\_Ref, ControlNet \cite{controlnet, ip}, FreeControl \cite{freecontrol}, Ctrl-X \cite{ctrlx}, Uni-ControlNet \cite{uni}, T2I-Adapter \cite{t2i-adapt, ip}, and Reimagine \cite{reimagine}. (b) The interface of our user study.}
  \label{fig:room}
\end{figure}

Note that because of the strong constraints from the image condition of TRELLIS \cite{trellis}, there is little diverse space for direct NOFT\_3D\_Img. Therefore, we first synthesize the image variants based on NOFT\_2D and then conduct 3D modeling based on the trellis-image-large model. Figure \ref{fig:sup_robot} shows the NOFT results. Text-based NOFT\_3D uses the trellis-text-xlarge model, as shown in Figure \ref{fig:3d_sup}. 
\paragraph{User Study}
We invite 10 users to conduct the subjective study. First, we briefly explain the highly correlated asset creation task. We suggest that users carefully observe the original content and generated image variants obtained by 6 state-of-the-art methods and our proposed NOFT. Each observed algorithm has 20 samples. These observers need to select the better image variant set from 3 aspects: (a) overall quality, (b) overall fidelity considering structure and appearance, (c) controllable diversity subject to the fidelity. The interface of our user study is shown in Figure \ref{fig:room}.



\end{document}